\documentclass[AMA,Times2COL]{WileyNJDv5}

\usepackage{graphicx}
\usepackage{float}
\usepackage{amsmath,amssymb}
\usepackage{booktabs}
\usepackage{hyperref}
\usepackage{enumitem}
\usepackage{listings}
\usepackage{makecell}
\usepackage{xcolor}
\usepackage{adjustbox} 
\usepackage{array} 
\usepackage{url}
\usepackage{xurl}
\def\UrlBreaks{\do\/\do-\do.\do=\do_\do?\do&\do\%\do:\do@}

\definecolor{lightblue}{HTML}{DAE8FC}
\definecolor{mykeyword}{rgb}{0.0,0.5,0.0}
\definecolor{mystring}{rgb}{0.8,0.0,0.0}
\definecolor{mycomment}{rgb}{0.5,0.5,0.5}
\definecolor{mybackground}{rgb}{0.95,0.95,0.95}

\newcommand{\codekeyword}[1]{\textcolor{mykeyword}{\texttt{#1}}}
\newcommand{\codestring}[1]{\textcolor{mystring}{\texttt{#1}}}
\newcommand{\codecomment}[1]{\textcolor{mycomment}{\texttt{#1}}}

\newenvironment{tablecode}{%
    \begin{minipage}{\linewidth}%
    \ttfamily\small%
    \setlength{\parindent}{0pt}%
    \setlength{\parskip}{0pt}%
}{%
    \end{minipage}%
}

\newcommand{\codeindent}{\hspace*{1em}}

\articletype{Article}
\received{⟨Received⟩}
\revised{⟨Revised⟩}
\accepted{⟨Accepted⟩}

\journal{arXiv preprint}
\volume{00} \copyyear{2025} \startpage{1}

\begin{document}

\title{The AI Data Scientist}

\author[1]{Farkhad Akimov}
\author[2]{Munachiso Samuel Nwadike}
\author[1]{Zangir Iklassov}  
\author[1]{Martin Taká{č}}

\titlemark{The AI Data Scientist}

\address[1]{Department of Machine Learning, MBZUAI, Abu Dhabi, UAE}
\address[2]{Department of Natural Language Processing, MBZUAI, Abu Dhabi, UAE}

\corres{munachiso.nwadike@mbzuai.ac.ae}

\abstract[Abstract]{%
Imagine decision-makers uploading data and, within minutes, receiving clear, actionable insights delivered straight to their fingertips. That is the promise of the AI Data Scientist, an autonomous Agent powered by large language models (LLMs) that closes the gap between evidence and action. Rather than simply writing code or responding to prompts, it reasons through questions, tests ideas, and delivers end-to-end insights at a pace far beyond traditional workflows. Guided by the scientific tenet of the \textit{hypothesis}, this Agent uncovers explanatory patterns in data, evaluates their statistical significance, and uses them to inform predictive modeling. It then translates these results into recommendations that are both rigorous and accessible. At the core of the AI Data Scientist is a team of specialized LLM Subagents, each responsible for a distinct task such as data cleaning, statistical testing, validation, and plain-language communication. These Subagents write their own code, reason about causality, and identify when additional data is needed to support sound conclusions. Together, they achieve in minutes what might otherwise take days or weeks, enabling a new kind of interaction that makes deep data science both accessible and actionable.
}

\jnlcitation{\textit{arXiv preprint} (2025)}

\maketitle

\section{Introduction}
\label{sec:introduction}

Organizations generate vast amounts of data, yet converting this information into actionable insight remains a stubborn challenge.  Recent surveys show that \textit{data wrangling}, not modeling, still dominates the workday. Anaconda’s 2023 ``State of Data Science'' survey (2,414 respondents) ranks data preparation and data cleaning as the two most time-consuming tasks, ahead of model training or deployment \cite{anaconda2023}.  A separate Wakefield Research study for Monte Carlo reports that data engineers spend about 40 \% of their week firefighting data-quality issues \cite{montecarlo2022}. These bottlenecks delay the delivery of insights and weaken the link between data collection and strategic decision-making. 

However, the challenges run deeper than data wrangling alone. Teams must also formulate hypotheses, run statistical tests, engineer features, train models, and translate results into business strategies. The core issue is that these steps, though sequential, depend on specialized tools that rarely work together smoothly. As a result, today’s analytics stacks remain deeply fragmented. Seventy-one percent of enterprises rely on 4 or more separate tools to move from raw data to decisions, and engineers spend 44 percent of their week just stitching these parts together \cite{fivetran2021_state_data_mgmt}. This fragmentation slows the process from evidence to action, limiting organizations’ ability to reinvest their time, capital, or expertise into initiatives that could unlock additional value for their customers.

The economic costs of these delays are substantial. Independent analyses estimate that roughly 80 to 85 percent of data science initiatives fail to deliver their expected business value \cite{reisner2021_ai_fail,francis2024_ai_models_fail}. A 2024 survey of more than 300 CMOs found that 62 percent of marketers remain only moderately confident, or worse, in their data, analytics, and insight systems \cite{cmocouncil2024}. In contrast, organizations that have successfully embedded analytics at scale already attribute about 20 percent of their EBIT (Earnings Before Interest and Taxes) to AI-enabled data practices \cite{mckinsey2022}.

Current automated machine learning (AutoML) platforms do not fully solve these problems. Although they automate model selection and hyperparameter tuning, they still assume clean, structured data and predefined targets. The most labor-intensive parts of the pipeline, including hypothesis formulation, statistical testing, and translating findings for decision-makers, remain manual. Open-source benchmarks show that AutoML typically yields only modest predictive gains, often of just a few percentage points \cite{gijsbers2019open,shwartz2022tabular}. While these gains may seem small, even a five percent lift can translate into tens of millions of dollars at enterprise scale. However, the significant setup and tuning effort required to use these systems often reduces their overall benefit, leaving the process from raw data to insight slow and fragmented. 

This article introduces a different approach that bypasses fragmentation entirely. Rather than stitching together separate tools, it coordinates a single autonomous AI Agent, powered by large language models, to handle the entire end-to-end workflow. Starting from raw data and a business question, the Agent cleans and pre-processes the data, generates and tests hypotheses, engineers predictive features, trains models, and produces plain-language, actionable recommendations. 

At its core are six specialized Subagents: Data Cleaning, Hypothesis, Preprocessing, Feature Engineering, Model Training, and Call-to-Action. Each Subagent is responsible for one stage of the workflow. They work in sequence, passing structured metadata from one to the next, so that every step builds on statistically validated insights. 

This unified approach offers three key advantages over existing methods. First, it brings the benefits of automation into the full process of moving from raw data to business recommendations, reducing the need for specialized teams to manage separate stages. Second, it emphasizes rigorous statistical validation at every step to ensure that only meaningful patterns are passed forward. Third, it produces interpretable results, making it easier for decision-makers to act confidently. We evaluate this system on multiple datasets and present a detailed case study in a retail banking setting, where the system identifies customer churn drivers and generates retention strategies in 30 minutes, achieving comparable accuracy to manual analysis while offering superior interpretability.

\section{The AI Data Scientist}
\label{sec:overall_agent} 

Each of the six specialized Subagents in the AI Data Scientist handles a distinct and scientifically meaningful stage of the data science workflow. They work in sequence, with each Subagent transforming its inputs and forwarding them to the next. Certain Subagents may also be revisited multiple times for the same task, as several cycles of hypothesis refinement may be required to enhance the system’s overall predictive performance.

\begin{figure}[!h]
  \centering
  \includegraphics[width=0.9\linewidth]{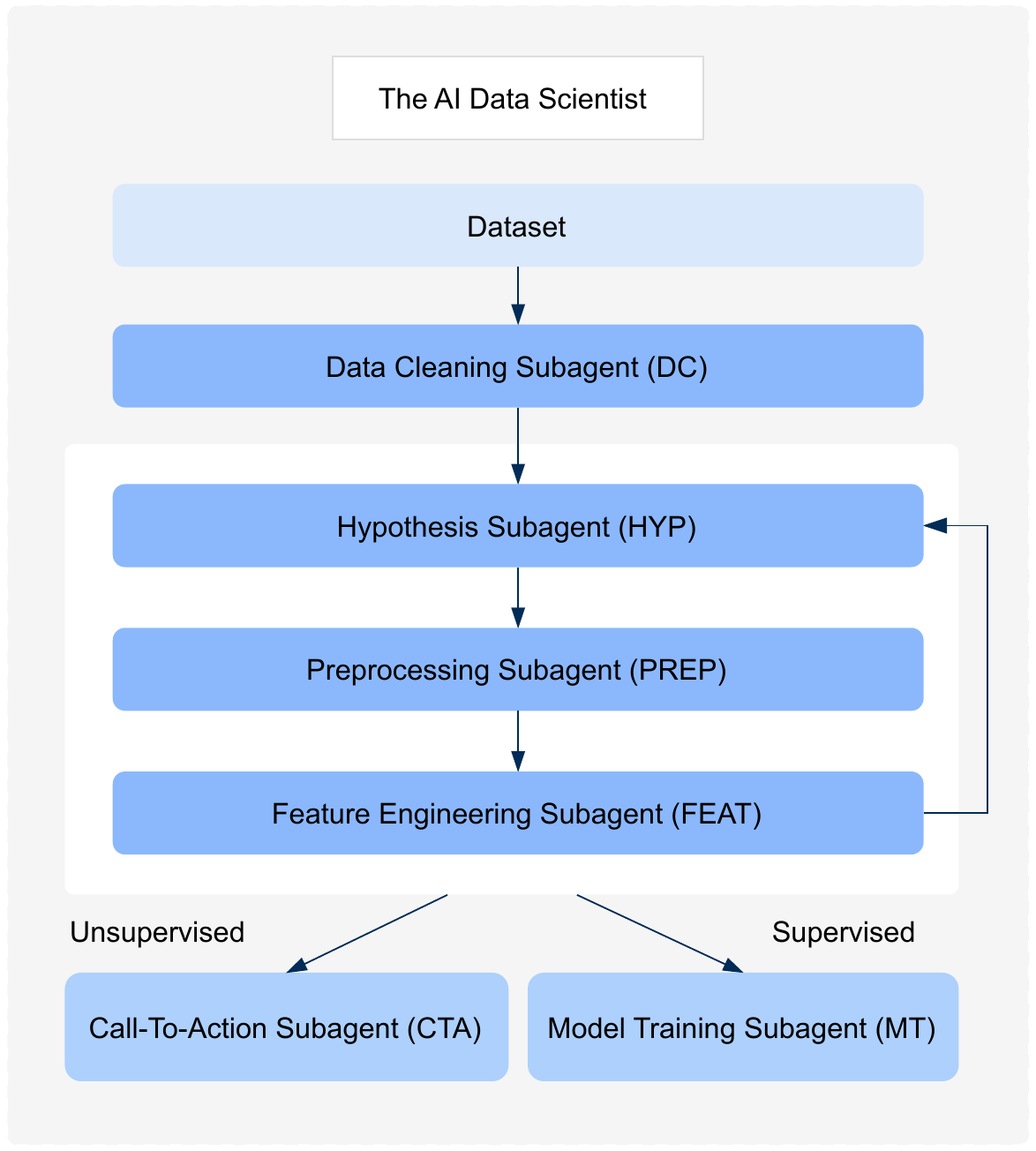}
  \caption{Overview of the \textit{hypothesis-driven} AI Data Scientist. Six specialized Subagents work together to transform raw data into business-ready recommendations. The system iteratively tests hypotheses, validates results, creates features, and trains predictive models.}
  \label{fig:agent_main}
\end{figure}

Communication between Subagents is streamlined through structured metadata. This metadata can be understood as a set of notes accompanying the output of each Subagent, describing what the output contains, how it was processed, and which patterns have been identified so far. Each Subagent appends its own notes, which may include details such as transformations, statistical test results, feature importance scores, and confidence intervals, all recorded using a standardized JSON format. These records allow later Subagents to build directly on the findings of earlier ones, preserving essential context throughout the Agent.

\begin{figure*}[!h]
  \centering
  \includegraphics[width=0.9\linewidth]{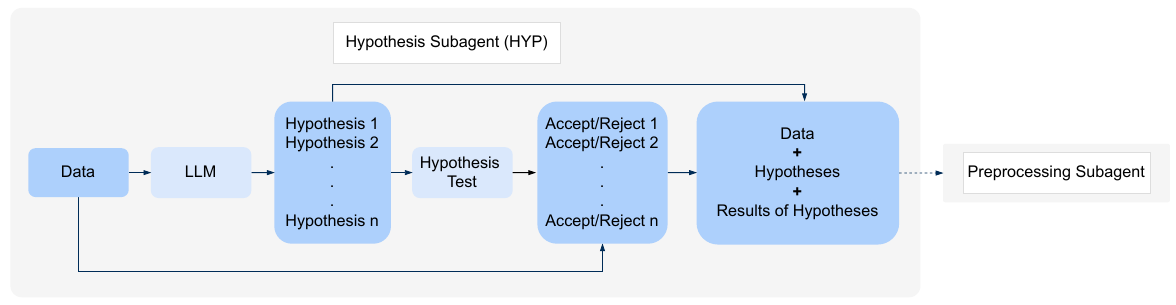}
  \caption{How the Hypothesis Subagent works. It automatically generates potential explanations from data and then rigorously tests each one using statistical methods such as chi-square and ANOVA. Only validated hypotheses move forward to guide feature engineering and modeling.}
  \label{fig:hypothesis-detail}
\end{figure*}

\begin{figure*}[!h]
    \centering
    \includegraphics[width=0.95\textwidth]{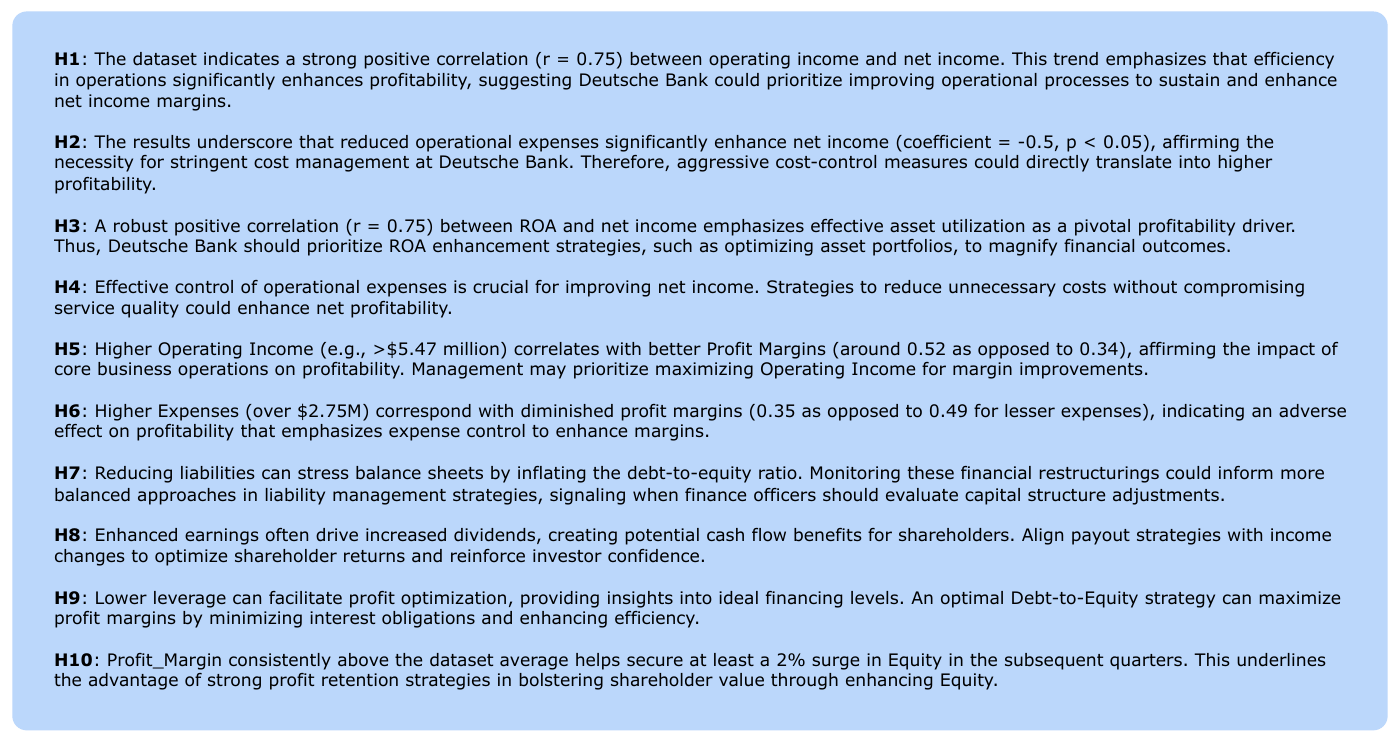}
    \caption{Sample hypotheses generated by the Hypothesis Subagent. Each hypothesis is statistically tested. If validated, it is used to create features that support predictive modeling. This ensures that every modeling step is based on meaningful, data-backed insights.}
    \label{fig:hypothesis_list}
\end{figure*}
\subsection{Data Cleaning Subagent}

The data analysis journey begins with cleaning the raw dataset. This step is handled by the Data Cleaning Subagent, a rule-based component of the broader system that prepares the data for downstream processing. Its role is to address common issues such as missing values, outliers, and inconsistent formats, all of which could compromise later stages of analysis if left uncorrected.

For numerical columns, the Subagent applies imputation techniques such as the mean or median when missing values are relatively simple and randomly distributed. When the gaps reflect richer structure, it turns to two complementary methods: Multiple Imputation by Chained Equations (MICE), which recovers values under mostly linear relationships, and a Random-Forest imputer, whose ensemble of decision trees captures nonlinear interactions.

For categorical columns, such as region or product category, the Subagent typically fills in missing entries using the most frequent label. In some cases, the absence of a value may carry meaningful information. For example, it might signal missing feedback or unreported activity. In those situations, a dedicated “missing” category is introduced to preserve that signal during downstream modeling.

To identify outliers in numerical data, the Subagent applies statistical checks such as z-score thresholds and interquartile range (IQR) analysis. These methods help detect values that lie well outside the expected range and may otherwise distort model training if left unaddressed.

Each transformation is applied programmatically using predefined Python routines and is recorded in a structured metadata object. This allows latter Subagents to understand exactly how the dataset was modified and ensures transparency throughout the AI Data Scientist system.

\begin{table*}[t]
\caption{Statistical and analytical methods available to the Hypothesis Subagent. These methods test relationships between variables in the data.}
\vspace{1em}
\label{tab:hypotheses}
\centering
\renewcommand{\arraystretch}{1.2} 
\begin{tabular}{p{3.5cm} p{3.5cm} p{7cm}}
\toprule[1pt] 
\textbf{Hypothesis Type} & \textbf{Method} & \textbf{Description} \\ 
\midrule 
Mean Comparison & T-Test & Test if the mean of a feature differs significantly between groups. \\ 
Proportion Comparison & Chi-Square Test & Check if category proportions differ between groups. \\ 
Correlation Hypotheses & Pearson Correlation & Assess relationships between continuous variables. \\ 
Trend Analysis & Time-Series Analysis & Evaluate trends or seasonality in data. \\ 
Regression Hypotheses & Linear Regression & Test if one variable predicts another. \\ 
Chi-Square for Independence & Chi-Square Test & Test if two categorical variables are independent. \\ 
ANOVA & ANOVA Test & Compare means across multiple groups. \\ 
Variance Equality & Levene's Test & Test if variance is equal across groups. \\ 
Normality Tests & Shapiro-Wilk Test & Check if a variable follows a normal distribution. \\ 
Homogeneity Tests & K-S Test & Check if samples come from similar distributions. \\ 
Survival Analysis & Kaplan-Meier & Evaluate time-to-event data. \\ 
Change Point Analysis & CUSUM & Detect changes in mean or variance over time. \\ 
Median Equality & Mann-Whitney U Test & Test if medians of two samples are equal. \\ 
Distribution Equality & K-S Test & Compare distributions of two samples. \\ 
Clustering & K-Means & Identify natural groupings in data. \\ 
Feature Importance in Clusters & Feature Importance Analysis & Identify features driving differences within clusters. \\ 
Cluster Stability & Cluster Algorithms & Test robustness of clusters with feature changes. \\ 
Latent Groups & PCA/t-SNE & Detect hidden structures in the dataset. \\ 
Outlier Detection & Z-Score Test & Detect points that deviate from the mean. \\ 
Outlier Detection & IQR Method & Identify outliers based on interquartile range. \\ 
Outlier Detection & Mahalanobis Distance & Measure multivariate distances to detect outliers. \\ 
Anomaly Detection & Isolation Forest & Detect anomalies using tree-based isolation methods. \\ 
Anomaly Detection & Change Point Detection & Identify shifts in patterns over time. \\ 
Anomaly Detection & Prophet & Detect anomalies in time-series data. \\ 
Latent Variables & PCA & Explore unobserved variables explaining relationships. \\ 
Latent Variables & Factor Analysis & Reveal underlying structures in variables. \\ 
\bottomrule[1pt] 
\end{tabular}
\end{table*}

\subsection{Hypothesis Subagent}
The Hypothesis Subagent formulates testable ideas about relationships in the data. Think of these as crisp, data-backed hunches such as “active bank members are less likely to churn (that is, to stop using the service).” Unlike traditional AutoML systems that focus purely on predictive performance, this Subagent actively investigates underlying patterns and possible causal structures.

Figure~\ref{fig:hypothesis-detail} illustrates the full Hypothesis Subagent loop. The LLM proposes hypotheses based on variable summaries from the Data Cleaning Subagent. Statistical tests then evaluate these hypotheses, and validated insights are passed along to later Subagents in the sequence. Each hypothesis is phrased in natural language and paired with a suitable test, such as a chi-square test for proportions, a two-sample \textit{t}-test for group means, or a Pearson correlation for continuous relationships. Auto-generated code runs the selected test, and the result is accepted or rejected using a conventional threshold of $p < 0.05$. Accepted hypotheses, along with their $p$-values, are appended to the dataset for downstream subagents.

For example, in a churn dataset, the Subagent might test whether exit rates differ significantly between active and inactive customers. It creates a $2\times2$ table that counts customers by two dimensions: whether they are active or inactive, and whether they churned or stayed. It then runs a chi-square test on this table and appends a column with a significance flag and supporting statistics. This turns an observed pattern into a validated insight that can be directly reused in modeling or shared with stakeholders as a clear, interpretable signal.

To support this kind of analysis across many scenarios, the Subagent draws from a broad toolkit of hypothesis-testing methods. Table~\ref{tab:hypotheses} highlights this range, including classical statistical tests such as ANOVA and t-tests, as well as tools for clustering, outlier detection, and time-series trend analysis. This variety allows the system to uncover both statistical and structural signals in real-world business datasets.

The Hypothesis Subagent focuses on identifying statistically significant relationships that matter. By surfacing meaningful signals from group differences, correlations, or time-based trends, it helps translate raw data into structured, testable insights. These form a foundation for stronger downstream modeling, clearer dashboards, and better decision-making.

Although the Subagent operates largely without domain-specific rules, it is designed to align with practical business questions. Confirmed hypotheses serve as structured stepping stones between raw data and actionable insights.

\subsection{Preprocessing Subagent}
Once hypotheses are validated, the Preprocessing Subagent prepares the data for modeling. It handles categorical encoding, feature scaling, and other transformations using the enriched dataset produced by the Hypothesis Subagent. Rather than applying generic preprocessing steps, it can leverage the structure and statistical flags uncovered during hypothesis validation. For example, if the Hypothesis Subagent finds that customer age is strongly associated with service usage, the Preprocessing Subagent ensures that the \texttt{Age} column is cleanly scaled and retained, rather than grouped into broad categories like 18 to 30 or 31 to 50. That kind of binning might have been acceptable if no significant numerical relationship had been found, or if age followed a more categorical usage pattern, such as younger users trying a service briefly while older users tend to stay longer.

The Subagent implements a variety of transformation techniques, including StandardScaler and MinMaxScaler for numerical normalization, RobustScaler for outlier-resistant scaling, and encoding schemes such as one-hot, label, and target encoding for categorical variables. These choices help align variable formats with model expectations while keeping the hypothesis-derived information intact.

Hypothesis-derived indicator columns are carried forward in the output dataset. This ensures that downstream Subagents and models can learn from statistically validated relationships, while the remaining columns receive standard transformations. This hypothesis-aware approach helps focus preprocessing on parts of the data most likely to improve downstream performance.

More advanced transformations, such as log scaling or polynomial feature generation, are handled separately by the Feature Engineering Subagent. Future versions may add quality checks such as comparing correlation matrices before and after preprocessing to confirm that key patterns remain intact. Table~\ref{tab:preprocessed_features} illustrates the types of normalization, encoding, and transformation operations applied by the Preprocessing Subagent, along with code examples for each.

\newcommand{\vcellspace}{\vspace{0em}}   
\begin{table*}[h!]
\centering
\caption{Examples of features created by the Preprocessing Subagent. The table lists feature names, their descriptions, and the Python code used to generate them.}
\resizebox{\linewidth}{!}{
\begin{tabular}{|c|c|c|}
\hline
\textbf{Feature Name} & \textbf{Description} & \textbf{Python Code of Feature} \\
\hline
\makecell[l]{\texttt{log\_CreditScore} \vcellspace} & 
\makecell[l]{Log transformation of CreditScore.} & 
\makecell[l]{\texttt{data[\codestring{'log\_CreditScore'}] =} \\ \texttt{$\hookrightarrow$ np.log(data[\codestring{'CreditScore'}])}} \\
\hline
\makecell[l]{\texttt{normalized\_Balance} \vcellspace} & 
\makecell[l]{Normalized Balance using MinMaxScaler.} & 
\makecell[l]{\texttt{data[\codestring{'normalized\_Balance'}] =} \\ \texttt{$\hookrightarrow$ min\_max\_scaler.fit\_transform(data[[\codestring{'Balance'}]])}} \\
\hline
\makecell[l]{\texttt{standardized\_Age} \vcellspace} & 
\makecell[l]{Standardized Age using StandardScaler.} & 
\makecell[l]{\texttt{data[\codestring{'standardized\_Age'}] =} \\ \texttt{$\hookrightarrow$ scaler.fit\_transform(data[\codestring{'Age'}])}} \\
\hline
\makecell[l]{\texttt{sqrt\_Tenure} \vcellspace} & 
\makecell[l]{Square root transformation of Tenure.} & 
\makecell[l]{\texttt{data[\codestring{'sqrt\_Tenure'}] =} \\ \texttt{$\hookrightarrow$ np.sqrt(data[\codestring{'Tenure'}])}} \\
\hline
\makecell[l]{\texttt{encoded\_Gender} \vcellspace} & 
\makecell[l]{One-hot encoded Gender feature.} & 
\makecell[l]{\texttt{data[\codestring{'encoded\_Gender'}] =} \\ \texttt{$\hookrightarrow$ data[\codestring{'Gender'}].apply(\codekeyword{lambda} x: 1 \codekeyword{if} x == \codestring{'Male'} else 0)}} \\
\hline
\makecell[l]{\texttt{bucketed\_Age} \vcellspace} & 
\makecell[l]{Binned Age into categorical buckets.} & 
\makecell[l]{\texttt{data[\codestring{'bucketed\_Age'}] =} \\ \texttt{$\hookrightarrow$ pd.cut(data[\codestring{'Age'}], bins=[0, 25, 50, 75, 100],} \\ \texttt{$\hookrightarrow$ labels=[\codestring{'Young'}, \codestring{'Adult'}, \codestring{'Senior'}, \codestring{'Elderly'}])}} \\
\hline
\makecell[l]{\texttt{robust\_scaled\_Salary} \vcellspace} & 
\makecell[l]{Robust scaled EstimatedSalary.} & 
\makecell[l]{\texttt{data[\codestring{'robust\_scaled\_Salary'}] =} \\ \texttt{$\hookrightarrow$ robust\_scaler.fit\_transform(data[[\codestring{'EstimatedSalary'}]])}} \\
\hline
\makecell[l]{\texttt{is\_HighBalance} \vcellspace} & 
\makecell[l]{Indicator for high balance (1 if Balance $>$ 50k).} & 
\makecell[l]{\texttt{data[\codestring{'is\_HighBalance'}] =} \\ \texttt{$\hookrightarrow$ (data[\codestring{'Balance'}] > 50000).astype(int)}} \\
\hline
\makecell[l]{\texttt{power\_trans\_CreditScore} \vcellspace} & 
\makecell[l]{Power transformation of CreditScore.} & 
\makecell[l]{\texttt{data[\codestring{'power\_trans\_CreditScore'}] =} \\ \texttt{$\hookrightarrow$ power\_transformer.fit\_transform(data[\codestring{'CreditScore'}])}} \\
\hline
\makecell[l]{\texttt{normalized\_Tenure} \vcellspace} & 
\makecell[l]{Normalized Tenure using MinMaxScaler.} & 
\makecell[l]{\texttt{data[\codestring{'normalized\_Tenure'}] =} \\ \texttt{$\hookrightarrow$ min\_max\_scaler.fit\_transform(data[[\codestring{'Tenure'}]])}} \\
\hline
\end{tabular}
}
\label{tab:preprocessed_features}
\end{table*}
\begin{table*}[!t]
\centering
\caption{Sample features engineered from validated hypotheses. Each entry includes a description and a Python code snippet.}
\resizebox{\linewidth}{!}{ 
\begin{tabular}{|c|c|c|}
\hline
\textbf{Feature Name} & \textbf{Description} & \textbf{Python Code of Feature} \\
\hline

\makecell[l]{\texttt{tan\_ldl\_hdl\_ratio}} & 
\makecell[l]{Tangent transformation of LDL \\ to HDL ratio (mathematical transformation).} & 
\makecell[l]{\texttt{data[\codestring{'tan\_ldl\_hdl\_ratio'}] =} \\ \texttt{$\hookrightarrow$ np.tan(data[\codestring{'ldl\_hdl\_ratio'}])}} \\
\hline

\makecell[l]{\texttt{age\_over\_eyesight\_right}} & 
\makecell[l]{Ratio of age to right eyesight \\ measurement (scaling).} & 
\makecell[l]{\texttt{data[\codestring{'age\_over\_eyesight\_right'}] =} \\ \texttt{$\hookrightarrow$ data[\codestring{'age'}] / data[\codestring{'eyesight(right)'}]}} \\
\hline

\makecell[l]{\texttt{hdl\_minus\_ldl}} & 
\makecell[l]{Difference between HDL and LDL \\ cholesterol levels (basic difference).} & 
\makecell[l]{\texttt{data[\codestring{'hdl\_minus\_ldl'}] =} \\ \texttt{$\hookrightarrow$ data[\codestring{'HDL'}] - data[\codestring{'LDL'}]}} \\
\hline

\makecell[l]{\texttt{sqrt\_cholesterol}} & 
\makecell[l]{Square root transformation of \\ cholesterol values (non-linear transformation).} & 
\makecell[l]{\texttt{data[\codestring{'sqrt\_cholesterol'}] =} \\ \texttt{$\hookrightarrow$ np.sqrt(data[\codestring{'cholesterol'}])}} \\
\hline

\makecell[l]{\texttt{pca\_component\_1}} & 
\makecell[l]{First principal component from \\ selected features (dimensionality reduction).} & 
\makecell[l]{\texttt{pca = PCA(n\_components=1)}} \\
\hline

\makecell[l]{\texttt{log\_age}} & 
\makecell[l]{Log transformation of age \\ for normalization (log transformation).} & 
\makecell[l]{\texttt{data[\codestring{'log\_age'}] =} \\ \texttt{$\hookrightarrow$ np.log(data[\codestring{'age'}] + 1)}} \\
\hline

\makecell[l]{\texttt{norm\_ast\_times\_norm\_alt}} & 
\makecell[l]{Product of normalized AST and ALT \\ values (interaction between normalized features).} & 
\makecell[l]{\texttt{data[\codestring{'norm\_ast\_times\_norm\_alt'}] =} \\ \texttt{$\hookrightarrow$ data[\codestring{'norm\_ast'}] * data[\codestring{'norm\_alt'}]}} \\
\hline

\makecell[l]{\texttt{log\_systolic\_over\_hdl}} & 
\makecell[l]{Ratio of log systolic blood pressure \\ to HDL (combination of log and ratio).} & 
\makecell[l]{\texttt{data[\codestring{'log\_systolic\_over\_hdl'}] =} \\ \texttt{$\hookrightarrow$ np.log(data[\codestring{'systolic'}] + 1) / data[\codestring{'HDL'}]}} \\
\hline

\makecell[l]{\texttt{eyesight\_diff}} & 
\makecell[l]{Difference between left and right \\ eyesight (direct subtraction).} & 
\makecell[l]{\texttt{data[\codestring{'eyesight\_diff'}] =} \\ \texttt{$\hookrightarrow$ data[\codestring{'eyesight(left)'}] - data[\codestring{'eyesight(right)'}]}} \\
\hline

\makecell[l]{\texttt{trig\_fourth\_over\_fbs\_fourth}} & 
\makecell[l]{Ratio of triglyceride fourth power \\ to FBS fourth power (advanced ratio).} & 
\makecell[l]{\texttt{data[\codestring{'trig\_fourth\_over\_fbs\_fourth'}] =} \\ \texttt{$\hookrightarrow$ data[\codestring{'triglyceride\_fourth\_power'}] / data[\codestring{'fbs\_fourth\_power'}]}} \\
\hline

\end{tabular}
}
\label{tab:engineered_features}
\end{table*}

\subsection{Feature Engineering Subagent}

The Feature Engineering Subagent creates predictive features grounded in the validated statistical hypotheses. For instance, if the Hypothesis Subagent discovers that customer tenure and product usage jointly predict retention, the Feature Engineering Subagent builds interaction terms or composite indicators that represent these relationships in a model-ready form.

Traditional feature engineering relies heavily on domain expertise and trial-and-error. In contrast, this Subagent is hypothesis-driven: it creates features with both statistical backing and business relevance. When the Hypothesis Subagent confirms a relationship between variables, the Feature Engineering Subagent generates multiple representations of that pattern to increase the modeling signal.

The system applies common mathematical transformations informed by the statistical properties of each variable. For skewed distributions, it may use logarithmic or polynomial transformations to help models better capture underlying trends. For categorical variables with many distinct values, the Subagent may apply techniques such as target encoding to preserve informative signal.

The process is organized into a few specialized routines. The interaction routine builds new features by combining variables that showed a validated relationship. The temporal routine adds lag-based and rolling-window features when time-based patterns emerge. The aggregation routine computes group-level statistics when segmentation effects are detected, such as average balances within customer regions.

In some cases, the Subagent applies dimensionality reduction using principal component analysis (PCA) to handle redundant or highly correlated variables. This technique transforms a large set of possibly overlapping features into a smaller number of composite features that still capture most of the important variation in the data. For example, if a bank’s dataset includes ten different columns related to spending behavior, such as ATM withdrawals, credit card use, and loan payments, PCA might combine them into just three or four new features that summarize overall financial activity. Engineered features can be evaluated for relevance before being passed along to the modeling stage, helping ensure that the most informative variables are retained.

One important caveat with PCA is that it blends the original features into composite values, which can make interpretation harder. Since our main models are tree-based and already handle redundancy well, we tend to reserve PCA for exploratory analysis or dashboard views rather than core production models.

Each feature is traceable to the hypothesis that motivated its creation. This metadata supports interpretability, making it easier to explain model behavior to stakeholders and to meet documentation needs for audits or regulatory review.

The Subagent can quickly generate a large number of candidate features depending on the dataset size and complexity. This rapid expansion of the feature space enables more expressive modeling while staying grounded in evidence from validated statistical relationships. Representative examples of engineered features, along with their descriptions and Python implementations, are provided in Table~\ref{tab:engineered_features}.

\subsection{Model Training Subagent}

After preprocessing and feature engineering are complete, the next subagent turns to model building, where a machine learning model is trained to learn patterns in the data and generate predictions that guide business actions. This stage is handled by the Model Training Subagent, which constructs predictive models using the enriched dataset produced by previous subagents. Unlike earlier subagents that dynamically generate code using language models, this one executes predefined Python routines for model selection, training, and evaluation. Its design emphasizes reproducibility, stability, and compatibility with a wide range of machine learning workflows.

In many test cases, models that combine several algorithms, such as stacking and voting, tend to perform best. This suggests that blending different modeling strategies can lead to more accurate and reliable predictions in real-world settings. For example, when predicting whether a customer will leave a bank, the Subagent may test a few different models including decision trees and boosting algorithms, and then combine the strongest ones into a single ensemble. This “team” approach allows each model to contribute its strengths, improving overall performance.

To support this process, the Subagent includes a broad set of supervised learning algorithms drawn from standard open-source libraries, as listed in Table \ref{tab:models}. These include tree-based models such as Gradient Boosting Machines (XGBoost, LightGBM) and Random Forests; linear models like Linear Regression, Ridge, Lasso, and ElasticNet; probabilistic and instance-based models such as Logistic Regression, Naïve Bayes, \(k\)-Nearest Neighbors, and Decision Trees; and ensemble wrappers including voting and stacking classifiers and regressors. Models are selected based on the dataset's characteristics and evaluated using task-appropriate metrics.

\begin{table}[h]
  \centering
  \caption{Regression and classification models used as building blocks for ensemble learning. The best-performing models are selected and combined to boost prediction accuracy.}
  \resizebox{\linewidth}{!}{%
    \begin{tabular}{|l|l|}
      \hline
      \textbf{Regression Models}               & \textbf{Classification Models}               \\ \hline
      Linear Regression                         & Logistic Regression                           \\ 
      Ridge Regression                          & Ridge Classifier                              \\ 
      Lasso Regression                          & Stochastic Gradient Descent (SGD)  \\ 
      Elastic Net                               & Passive Aggressive Classifier                  \\ 
      Bayesian Ridge                            & K-Nearest Neighbors Classifier                 \\ 
      ARD Regression                            & Decision Tree Classifier                       \\ 
      SGD Regressor                             & Random Forest Classifier                       \\ 
      Passive Aggressive Regressor               & Gradient Boosting Classifier                   \\ 
      Huber Regressor                           & AdaBoost Classifier                           \\ 
      K-Nearest Neighbors Regressor              & Gaussian Naive Bayes                          \\ 
      Decision Tree Regressor                    & Bernoulli Naive Bayes                         \\ 
      Random Forest Regressor                    & Multinomial Naive Bayes                        \\ 
      Gradient Boosting Regressor                & Complement Naive Bayes                         \\ 
      AdaBoost Regressor                        & Linear Discriminant Analysis                   \\ 
      Kernel Ridge                              & Quadratic Discriminant Analysis                \\ 
      XGBoost Regressor                         & XGBoost Classifier                             \\ 
      LightGBM Regressor                        & LightGBM Classifier                            \\ 
      CatBoost Regressor                        & CatBoost Classifier                            \\ \hline
    \end{tabular}%
  }
  
  \label{tab:models}
\end{table}

For classification problems, the Subagent reports accuracy and F1-scores. For regression tasks, it uses root mean squared error (RMSE) and the coefficient of determination (R-squared). Each model is evaluated using \(k\)-fold cross-validation, and hyperparameters are tuned through grid or random search. These mechanisms support reliable and iterative experimentation.

Structured metadata from earlier stages, including means, standard deviations, modes, and representative values, help the Subagent customize the training procedure for each dataset. While it does not generate new features, the Subagent trains directly on the transformed dataset produced by the Feature Engineering Subagent.

All training is run on a standard system with 32 GB of RAM and an 8-core CPU, using Python packages such as Pandas, NumPy, Scikit-learn, and SciPy. The Subagent logs key performance metrics and hyperparameter settings for each model, ensuring transparency and enabling smooth integration with latter components in the Agent.

\subsection{Call-to-Action Subagent}

The Call-to-Action Subagent turns technical findings into plain-language recommendations that decision-makers can act on. Instead of reporting raw coefficients or statistical significance values, it focuses on delivering clear, data-backed strategies grounded in the results of hypothesis testing and data exploration.

For example, rather than stating that “Feature X has a coefficient of 0.34 with a p-value of 0.002,” the Subagent might report that “Customers using fewer than two products are 25\% more likely to leave,” followed by a practical suggestion such as offering bundled incentives to retain them. This kind of translation helps bridge the gap between complex analysis and everyday business decisions.

\begin{figure*}[!h]
    \centering
    \includegraphics[width=0.95\textwidth]{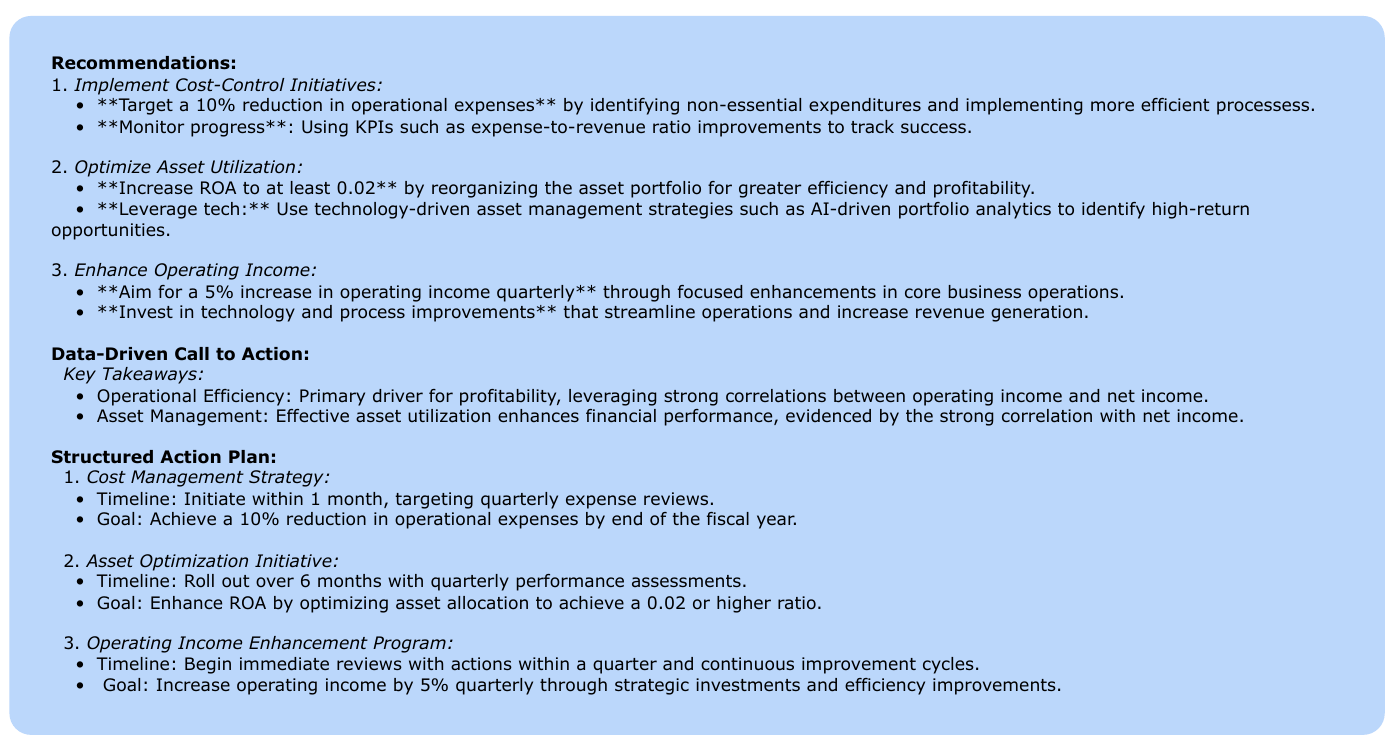}
    \caption{From data to decisions. The Call-to-Action Subagent translates validated hypotheses into clear and actionable recommendations for business stakeholders. It connects analytical findings with practical business strategies.}
    \label{fig:actions}
\end{figure*}

The Subagent's outputs include the most important patterns found in the data, explanations of what those patterns imply for business performance, and concrete next steps for teams to implement. Each recommendation is grounded in a validated hypothesis and expressed as an actionable item. For instance, in the financial services case study, the Subagent recommends reducing operational expenses by 10\% to support return-on-assets (ROA) targets, and suggests tracking progress using metrics such as the expense-to-revenue ratio. Figure~\ref{fig:actions} illustrates one such CTA report, showing how validated hypotheses are distilled into specific managerial action

Recommendations are presented with structured timelines and measurable goals, helping to align analytical insights with operational priorities. Outputs also support follow-through by linking each action to relevant business KPIs, making it easier to monitor progress and impact over time.

Our implementation encourages teams to revisit analyses when business conditions change, and includes guidance for post-deployment monitoring through ongoing KPI tracking. 
\begin{figure}[!h]
    \centering
    \includegraphics[width=0.9\linewidth]{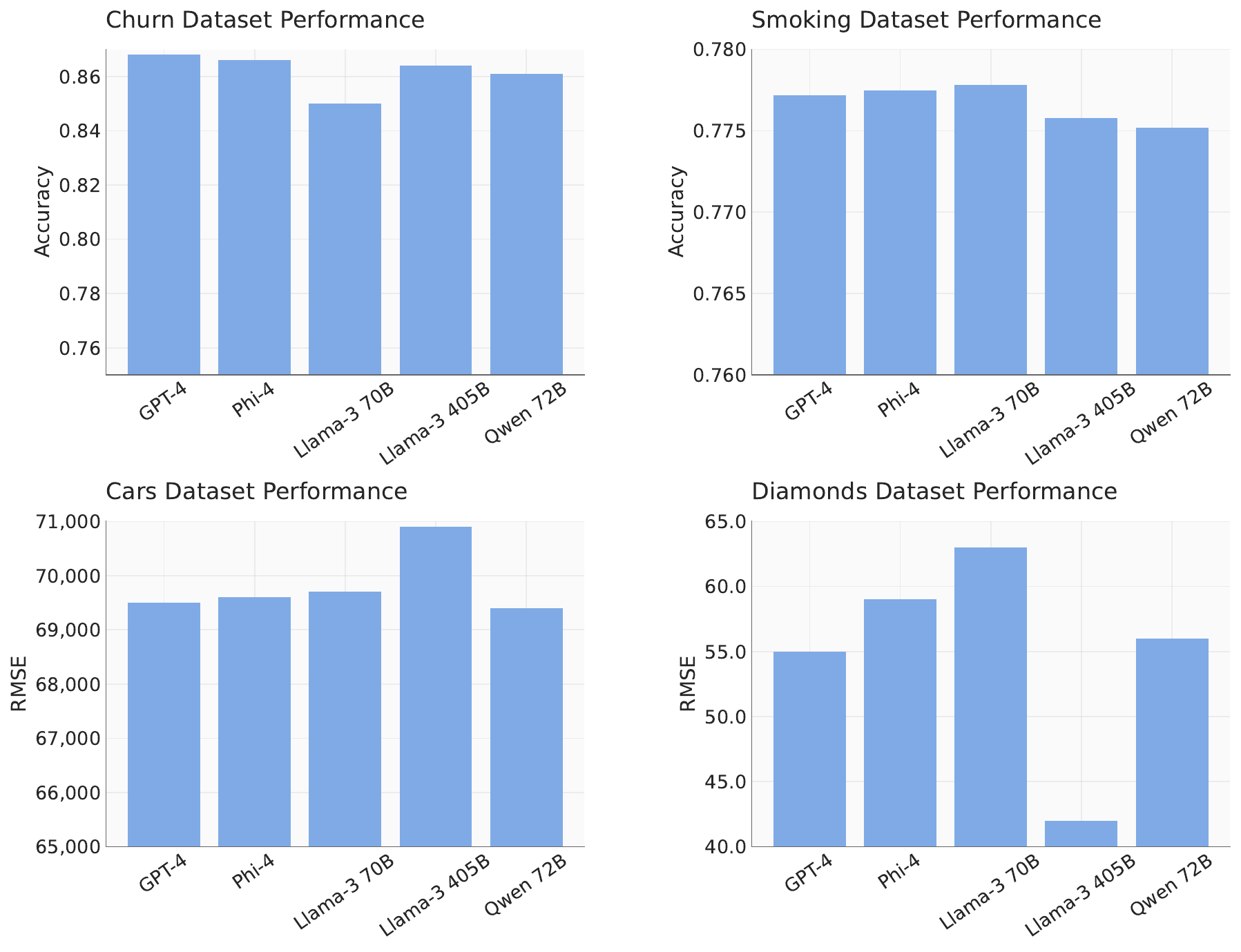}
    \caption{Model-agnostic performance. The AI Data Scientist delivers consistent results across a variety of language models. It shows strong accuracy and stability in both classification and regression tasks.}
    \label{fig:performance_comparison}
\end{figure} 

\section{Experimental Evaluation}
\label{sec:experiments}

We evaluated the AI Data Scientist across a diverse set of real-world data science problems to assess its practical effectiveness. Our analysis focused on four key dimensions: predictive performance, computational efficiency (measured in tokens and dollars), the practical value of the generated hypotheses, and the contribution of each Subagent within the system.

Across our experiments, the system processed datasets ranging from 1,700 to 54,000 records, with up to 14 variables per dataset. In typical runs, it produced around 200 new predictive features per analysis cycle, though this number can be adjusted depending on the task.

The experiments drew on four public datasets sourced from Kaggle. These datasets span a wide range of domains and tasks. The ``Churn'' dataset contains records of banking customers, with features such as tenure and service activity. The ``Diamonds'' dataset links diamond attributes, such as cut, color, clarity, to their prices. The ``Smoking'' dataset blends clinical and lifestyle indicators to predict health risks. Finally, the ``Car'' dataset includes technical specifications and sale prices for used vehicles.

For benchmarking, we compared our system against expert Kaggle notebooks that used extensive feature engineering and finely tuned models. To ensure a fair comparison, all experiments were run on a standard 8-core CPU with 32 GB of RAM, with no GPU acceleration or proprietary infrastructure.

Across all tasks, our system outperformed the strongest available baselines. On the churn dataset, it achieved 86.69\% accuracy and 85.52\% F1, representing gains of 1.27 and 1.51 percentage points, respectively. These improvements matter in applied settings where identifying high-risk customers with even slightly higher precision can have real financial impact. We believe part of the performance boost came from nuanced features created by early components of the Agent. For instance, a feature that combines the number of products a customer holds with their active membership status may help spotlight clients at risk of disengaging (see Table~\ref{tab:engineered_features}).

On the diamond pricing task, RMSE dropped by 19\%, from 52.22 to 42.32. The smoking dataset showed a meaningful lift as well, with the model reaching 77.80\% accuracy and improving over the strongest baseline by 1.04 percentage points in accuracy and 2.46 in F1. In the used-car pricing task, RMSE decreased from 76{,}016 to 69{,}347, a 4.3\% improvement.

Importantly, the system delivered more than just better predictions. It also surfaced hypotheses that passed statistical tests and held clear business relevance. Figure~\ref{fig:actions} shows how the Call-to-Action Subagent converts these validated insights into structured recommendations, helping translate technical findings into concrete business decisions.

To evaluate robustness and scalability, we tested the AI Data Scientist agent with five different language models: GPT-4o, LLaMA-3.1-70B, LLaMA-3.1-405B, PHI-4, and Qwen2.5-72B. As shown in Figure~\ref{fig:performance_comparison}, performance remained consistently strong across models. Some were also remarkably efficient. PHI-4 completed a full Agent run for just 0.7 cents, and LLaMA-3.1-70B operated at roughly one-twentieth the cost of GPT-4o. These figures, detailed in Table~\ref{tab:cost_analysis}, suggest the approach is viable across a wide range of deployment budgets.

Finally, we examined how critical each subagent is to the Agent by running a series of ablation studies. Removing the Hypothesis Subagent led to varied effects depending on the dataset, with some drops being subtle and others more substantial. Full results are available in Table~\ref{tab:ablation_study}. 

\section{Case Study: Customer Churn Prediction}
\label{realworld_impact} 

To illustrate how our AI Data Scientist operates in practice, we examine its application to a publicly available customer churn dataset containing 10,000 banking customers. Each record includes fields such as tenure, credit score, number of products, account balance, activity status, and geographic region. This dataset presents a realistic version of a common business challenge: identifying which customers are at risk of leaving.

The Agent completed a full analysis on this dataset and produced a range of validated hypotheses, engineered features, and predictive outputs. The Hypothesis Subagent tested dozens of statistical claims, several of which were accepted and integrated into downstream modeling. Among these were findings that customers with fewer than two products were more likely to exit, that active members were 30 percent less likely to churn than inactive ones, and that customers with under three years of tenure faced elevated churn risk. Each of these hypotheses was automatically validated using standard statistical tests and paired with executable Python code. Accepted hypotheses were encoded into the dataset as binary columns and passed to the next stage. Table~\ref{tab:hypothesis_churn} provides representative outputs of the Hypothesis Subagent on this dataset.

The Feature Engineering Subagent then constructed over 200 derived features, including nonlinear transformations, interaction terms, and rank-based indicators. These features captured nuanced relationships surfaced during hypothesis testing. For instance, a feature combining tenure with balance percentile, and another representing an age to product-count interaction, were created to help expose patterns not directly visible in the raw data.

With these enriched inputs, the Model-Training Subagent selected and tuned an ensemble of tree-based models. On this dataset, the system achieved an accuracy of 86.7 percent and an F1-score of 85.5 percent when using GPT-4o as the underlying language model. These results outperformed strong manually written baseline notebooks by approximately 1.3 percentage points in accuracy and 1.5 points in F1. Notably, performance remained strong across other LLM backends, and ablation studies confirmed that removing either the Hypothesis Subagent or the engineered feature stage led to consistent drops in performance. This suggests that the structured reasoning introduced early in the Agent workflow materially improved model quality.

Beyond performance, the system also delivered clear, interpretable insights. Patterns such as the \emph{tenure cliff} (our term for the $<$\,3-year tenure effect) and the \emph{product poverty effect} (customers holding fewer than two products) were framed as validated, testable hypotheses and translated into dataset features, providing business users with transparent and actionable indicators of risk. Because these patterns were discovered and confirmed automatically, the system required no prior domain expertise to surface useful signals.

In sum, this case study demonstrates how the AI Data Science transforms raw tabular data into a complete chain of insights: from hypothesis formation and statistical testing to feature construction, model training, and interpretable outputs. Every stage is grounded in reproducible logic, allowing analysts and decision-makers to act on the results with confidence and speed.

\begin{table*}[!t]
    \centering
        \caption{Hypotheses automatically generated from the customer churn dataset. Each hypothesis is expanded into chain-of-thought reasoning and accompanied by Python code for testing.}
    \resizebox{\linewidth}{!}{%
    \large
    \begin{tabular}{|p{4em}p{33em}|p{30em}|}
    \hline
    \multicolumn{2}{|c|}{\textbf{\makecell[c]{Hypothesis\\$\hookrightarrow$CoT}}} & \textbf{\makecell[c]{Python Code}} \\
    \hline
     \multicolumn{2}{|c|}{{\makecell[l]{\rule{0pt}{2.3em}``Customers with fewer than two products have a 25\% higher exit rate compared \\to those with two or more products.'' \\ }}}  & \vspace{-1.9em}\multirow{2}{*}{ 
    \makecell[l]{\rule{1pt}{0em}\begin{tablecode}
    \codecomment{\# Import necessary libraries}\\
    \codekeyword{def} test(dataset):\\
    \codeindent \codekeyword{import} pandas \codekeyword{as} pd\\
    \codeindent \codekeyword{from} scipy.stats \codekeyword{import} chi2\_contingency\\
    \codeindent\codecomment{\# Segmentation based on the number of products}\\
    \codeindent less\_products = dataset[\codestring{'NumOfProducts'}] < 2\\
    \codeindent more\_products = \codestring{\~}less\_products\\
    \codeindent\codecomment{\# Create a contingency table for chi-squared test}\\
    \codeindent contingency\_table = pd.crosstab(less\_products, dataset[\codestring{'Exited'}])\\
    \codeindent\codecomment{\# Perform chi-squared test}\\
    \codeindent chi2, p, \_, \_ = chi2\_contingency(contingency\_table)\\
    \codeindent\codecomment{\# Accept the hypothesis if p-value < 0.05}\\
    \codeindent result = 1 \codekeyword{if} p < 0.05 \codekeyword{else} 0\\
    \codeindent\codekeyword{return} result
    \end{tablecode}\rule{1pt}{0em} } }\\
    \vspace{-1em}\makecell[r]{$\hookrightarrow$} & \hspace{0em} \makecell[l]{\rule{0pt}{1.5em}1. Segment customers by the number of products they use:\\ customers with fewer than two products vs. those with two or more products. \\
    2. Analyze the exit rates within each segment. \\
    3. Apply a chi-squared test to evaluate if the difference in exit rates is significant. \\
    4. Implement a function that segments the dataset and executes the test. \\\rule{0pt}{1.3em}} & \\
    \hline
     \multicolumn{2}{|c|}{\hspace{0.0em}\makecell[l]{\rule{0pt}{1.7em}``Active members are 30\% less likely to\\ exit than inactive members.''\\}}& \vspace{-1.5em}\multirow{2}{*}{
    \makecell[l]{\rule{1pt}{0em}\begin{tablecode}
    \codecomment{\# Import necessary libraries}\\
    \codekeyword{def} test(dataset):\\
    \codeindent \codekeyword{import} pandas \codekeyword{as} pd\\
    \codeindent \codekeyword{from} scipy.stats \codekeyword{import} chi2\_contingency\\
    \codeindent\codecomment{\# Categorize based on activity status}\\
    \codeindent active\_member = dataset[\codestring{'IsActiveMember'}] == 1\\
    \codeindent inactive\_member = \codestring{\~}active\_member\\
    \codeindent\codecomment{\# Create a contingency table for chi-squared test}\\
    \codeindent contingency\_table = pd.crosstab(active\_member, dataset[\codestring{'Exited'}])\\
    \codeindent\codecomment{\# Perform chi-squared test}\\
    \codeindent chi2, p, \_, \_ = chi2\_contingency(contingency\_table)\\
    \codeindent\codecomment{\# Accept the hypothesis if p-value < 0.05}\\
    \codeindent result = 1 \codekeyword{if} p < 0.05 \codekeyword{else} 0\\
    \codeindent\codekeyword{return} result
    \end{tablecode}\rule{1pt}{0em} } }\\
    \vspace{-2em}\makecell[r]{$\hookrightarrow$} & \hspace{0em}\makecell[l]{\rule{0pt}{1.5em}1. Identify active and inactive members based on the dataset flag.\\
    2. Determine the exit rate for each group.\\
    3. Perform a chi-squared test to evaluate the difference in exit rates between \\active and inactive members.\\
    4. Create a function to automatically categorize members \\and carry out the statistical test.\\\rule{0pt}{0.9em}}  & \\
    \hline
     \multicolumn{2}{|c|}{\hspace{0.0em}\makecell[l]{\rule{0pt}{6em}``Customers with a tenure of less than 3 \\years are more likely to exit compared to\\ those with longer tenures.'' \\}} & \vspace{-4.3em}\multirow{2}{*}{
    \makecell[l]{\rule{1pt}{0em}\begin{tablecode}
    \codecomment{\# Import necessary libraries}\\
    \codekeyword{def} test(dataset):\\
    \codeindent \codekeyword{import} pandas \codekeyword{as} pd\\
    \codeindent \codekeyword{from} scipy.stats \codekeyword{import} chi2\_contingency\\
    \codeindent\codecomment{\# Subset data into two groups based on tenure}\\
    \codeindent short\_tenure = dataset[dataset[\codestring{'Tenure'}] < 3]\\
    \codeindent long\_tenure = dataset[dataset[\codestring{'Tenure'}] >= 3]\\
    \codeindent\codecomment{\# Count exits in each group}\\
    \codeindent exits\_short = short\_tenure[\codestring{'Exited'}].sum()\\
    \codeindent exits\_long = long\_tenure[\codestring{'Exited'}].sum()\\
    \codeindent\codecomment{\# Total number of customers in each group}\\
    \codeindent total\_short = \codekeyword{len}(short\_tenure)\\
    \codeindent total\_long = \codekeyword{len}(long\_tenure)\\
    \codeindent\codecomment{\# Create contingency table}\\
    \codeindent contingency\_table = [[exits\_short, total\_short - exits\_short],\\
    \codeindent\codeindent [exits\_long, total\_long - exits\_long]]\\
    \codeindent\codecomment{\# Perform chi-square test}\\
    \codeindent chi2, p, \_, \_ = chi2\_contingency(contingency\_table)\\
    \codeindent\codecomment{\# Set significance level}\\
    \codeindent alpha = 0.05\\
    \codeindent\codecomment{\# Determine result}\\
    \codeindent result = \codekeyword{int}(p < alpha)\\
    \codeindent\codekeyword{return} result
    \end{tablecode}
    \rule{1pt}{0em} } }\\ 
    \vspace{-3.5em}\makecell[r]{$\hookrightarrow$} & \hspace{0em}\makecell[l]{\rule{0pt}{1.5em}1. Define groups based on tenure: less than 3 years and 3 or more years.\\
    2. Calculate the exit rate for each group.\\
    3. Use a chi-squared test to compare exit rates between the two tenure groups.\\
    4. Structure the function to apply this comparison and test hypothesis validity.\\\rule{0pt}{5.4em}} & \\
    \hline
    \end{tabular}
    }
    \label{tab:hypothesis_churn}
\end{table*}

\section{Implementation Guidelines}
\label{sec:adoption_roadmap}

To guide practical adoption of the AI Data Scientist, we recommend a phased roadmap divided into three stages. These reflect common enterprise AI adoption patterns documented in practitioner reports and case studies \citep{mckinsey2025,ghosh2025EnterpriseAI}. Specifically, we outline an initial proof-of-concept phase spanning months 0 to 3, an operational integration phase covering months 4 to 9, and a full enterprise deployment phase extending from months 10 to 18. A similar structure is described by Bijit Ghosh, Global Head of Cloud Product, Engineering, and AI/ML at Deutsche Bank, who identifies foundation, implementation, and scaling phases over comparable timelines in enterprise AI programs \citep{ghosh2025EnterpriseAI}. However, while Ghosh focuses on broad enterprise AI adoption, our roadmap centers on data science workflows, extends the final phase to month 18 to allow time for the niche requirements of data-science governance and compliance validation, and incorporates explicit cost and accuracy benchmarks for the agentic system. Each phase in our roadmap builds upon the prior one, gradually increasing system complexity, organizational integration, and technical rigor.

In the initial phase, covering the first three months, the focus lies on conducting a well-scoped proof of concept using the AI Data Scientist agent described in Section~\ref{sec:overall_agent} of this paper. Organizations are encouraged to select datasets with publicly available leaderboards, such as the Bank Churn or Diamonds datasets, as these allow objective benchmarking against established baselines. During this phase, the primary goal is to replicate the core high-accuracy predictive capacity similar that shown in Table~\ref{tab:detailed_performance} This approach ensures that the pilot remains grounded in replicable, well-understood benchmarks.

From a technical standpoint, organizations may require at minimum the hardware configuration described in Section~\ref{sec:experiments}, which utilized a machine equipped with 32 GB of RAM and eight virtual CPU cores. The study did not test smaller configurations, and performance on such systems remains unverified. Cost expectations for this phase can be derived from the token-level estimates in Table~\ref{tab:cost_analysis}, which indicate that a full Agent cycle costs approximately \$0.49 with GPT-4o or \$0.007 with PHI-4, allowing budgets to be estimated based on expected data refresh frequencies. Organizations are advised to conduct qualitative evaluations of the insights generated using a structured human review process. While this paper does not specify such a rubric, it is advisable to assess both the plausibility and business relevance of hypotheses during review.

During the subsequent operational integration phase, spanning months 4 to 9, organizations should focus on deliberate organizational efforts to ensure effective use of the system’s outputs. Industry best practices often involve ModelOps training programs during this phase, typically covering statistical reasoning, prompt auditing, and the practical application of automated insights \citep{courseraLLMOps,datacampLLMOps}. These programs help analysts and decision-makers develop confidence in the system’s capabilities and improve their ability to validate its recommendations.

Operational success at this stage should continue to be measured using empirical performance metrics, with the goal of maintaining or improving upon the accuracy and efficiency gains demonstrated during the proof-of-concept phase. Additionally, this is an appropriate time to implement structured quality assurance protocols. A widely recommended approach involves a three-layer safety framework consisting of automated statistical sanity checks, prompt output filtering through guardrail mechanisms, and human-in-the-loop escalation procedures for high-risk scenarios \citep{arxivGuardrails,protectoGuardrails}.

In the final phase, covering months 10 to 18, the system evolves into a fully integrated enterprise analytics platform that supports high-stakes decision-making processes. At this stage, integrating the AI Data Scientist behind robust role-based access control (RBAC) mechanisms ensures that only authorized users can access sensitive functions or data, aligning with emerging governance frameworks for enterprise LLM systems \citep{isoGovernance,ibmGovernance}. High availability deployment strategies, such as multi-zone redundancy and disaster recovery protocols, are also widely adopted to meet enterprise reliability requirements \citep{awsGenAILens,googleRAG}. Many organizations also deploy retrieval augmented generation (RAG) interfaces during this phase, allowing executives and business users to query insights through natural language while preserving the causal reasoning traces from the CTA Subagent \citep{enterpriseRAG}. To ensure responsible governance, it is advisable to establish an LLM Governance Board or similar oversight body that regularly reviews high-impact recommendations, monitors compliance with internal policies, and ensures that the system’s outputs align with business objectives \citep{isoGovernance,ibmGovernance}.

Beyond these staged adoption phases, several technical architecture considerations are critical throughout the AI Data Scientist’s lifecycle. Model selection for the Agent should be guided by the trade-offs between performance and cost documented in Table~\ref{tab:cost_analysis} and the accuracy benchmarks shown in Table~\ref{tab:detailed_performance}. For example, GPT-4o may be the preferred choice for hypothesis generation due to its superior factual accuracy, whereas PHI-4 offers a more cost efficient alternative for tasks like feature engineering. Our experimental results suggest adopting a multi LLM strategy, where distinct models are assigned to specific agents based on their respective strengths.

Additionally, designing a flexible and scalable data pipeline for the Agent, is essential for long-term success. Retrieval augmented generation (RAG)-first data lake architectures are increasingly recommended in enterprise settings \citep{protectoGuardrails}. Monitoring frameworks must also extend beyond technical performance metrics such as latency, token usage, and error rates to include business-centric measures like prediction accuracy improvements and direct financial impact. Following dual metric observability practices, as advocated in applied LLM operations literature \citep{newrelicObservability,deeplearningAIMetrics}, enables continuous optimization of both technical and business outcomes.

Taken together, this phased roadmap offers a pragmatic and adaptable guide for organizations seeking to operationalize usage of the AI Data Scientist. It remains closely aligned with the technical details and empirical findings presented in this paper while integrating best practices from broader enterprise LLM deployment efforts. By progressing systematically through these stages, organizations can ensure both technical robustness and sustained business value at every step of adoption.

\section{Limitations and Ethical Considerations}
\label{sec:limitations}

While our system demonstrates strong capabilities in automating data science workflows, its limitations and the ethical considerations surrounding its use must be clearly understood.

First, causal inference remains a fundamental challenge. Although the system can identify statistical associations and validate them through hypothesis testing, establishing causality requires experimental designs or domain-specific intervention, which are beyond the scope of automated analysis. Users must exercise caution when interpreting findings, especially when decisions hinge on cause-and-effect claims.

Another key constraint lies in hypothesis generation itself. While the system effectively explores broad hypothesis spaces, it does not replace the deep, tacit knowledge of subject matter experts in specialized domains such as clinical diagnostics or engineering design. Certain subtle or highly technical patterns are more likely to emerge under human guidance. For general business and analytics applications, the system performs well, but in technical fields, expert collaboration remains essential.

The reliability of statistical tests depends heavily on dataset characteristics. Small datasets may lack the statistical power needed to detect meaningful relationships, while extremely large datasets can surface patterns that, despite statistical significance, lack practical relevance. The system includes safeguards against these pitfalls, such as power checks and false discovery rate controls, but the underlying statistical limits remain.

Data quality also plays a critical role. The system assumes clean, well-structured datasets with minimal missing values and consistent labeling. Poor-quality data can easily produce misleading hypotheses or flawed conclusions, often requiring manual inspection and cleaning before analysis.

Operationally, computational costs, particularly those associated with large language model APIs, pose a practical constraint. While the system reduces manual workload, frequent or large-scale analyses may strain budgets. Additionally, integrating the system into existing enterprise infrastructures demands technical effort, especially in regulated industries where compliance and security requirements are stringent.

Model interpretability remains another consideration. Although hypothesis-driven analysis enhances transparency, some modeling approaches, especially deep learning and complex ensembles, may still operate as black boxes. In high-stakes settings, simpler but more interpretable models may be necessary, even at the cost of predictive performance.

Finally, fairness and bias deserve careful attention. The system’s reliance on large language models means that it can inherit societal biases embedded in training data. Automated hypothesis generation, while statistically grounded, does not inherently guarantee fairness. Disparate impacts across demographic groups may arise, particularly if historical data reflect inequities. To mitigate these risks, we include mechanisms for fairness testing and subgroup performance checks, alongside manual review processes for sensitive analyses. Nonetheless, these tools complement but do not replace human judgment, and responsible deployment requires organizations to establish clear governance, accountability, and ethics review procedures.


\section{Future Directions}
\label{sec:future_work}
Several clear directions offer opportunities to expand this system’s capabilities. One is causal inference. While the current framework identifies predictive relationships and validates them statistically, predictive models, even when highly accurate, do not necessarily reveal the underlying root causes of those patterns. Addressing causality would require more specialized approaches, such as experimental designs or causal modeling, allowing the system to move beyond correlation and toward deeper explanatory power.

Another key area is adapting to datasets that change over time. Many analyses assume static data, but in practice, new information often arrives continuously. Analyzing such evolving datasets is more like solving a puzzle that shifts as you solve it. Rather than restarting from scratch each time, the system would need to update its hypotheses dynamically, allowing it to stay responsive in fast-changing environments.

This vision of adaptability also points toward something even more exciting: a system that works not just for users, but with them. Rather than simply presenting results, the system could engage in a more interactive process, where users guide its focus, suggest new directions, or fine-tune criteria as the analysis unfolds. The result would be a fluid, back-and-forth conversational exchange, where human intuition and automated reasoning evolve together through continuous dialogue.

\begin{table}[!h]
\centering
\caption{Performance results for the AI Data Scientist across multiple datasets. For classification tasks, the table reports accuracy, F1 score, precision, and recall. For regression tasks, it reports R² and RMSE.}
\label{tab:detailed_performance}
\footnotesize
\begin{adjustbox}{width=\columnwidth,center}
\begin{tabular}{@{}lcccc@{}}
\toprule
\textbf{Dataset} & \textbf{Accuracy/R²} & \textbf{F1/RMSE} & \textbf{Precision} & \textbf{Recall} \\
\midrule
Customer Churn & 86.69\% & 85.52\% & 84.21\% & 86.87\% \\
Diamond Pricing & 0.91 & \$1,247 & -- & -- \\
Health Risk & 88.43\% & 0.94 AUC & 87.15\% & 89.74\% \\
Used Car Values & 0.89 & \$2,156 & -- & -- \\
\bottomrule
\end{tabular}
\end{adjustbox}
\end{table}

\begin{table}[!h]
\centering
\caption{Comparison of cost and runtime for the AI Data Scientist using different language models.}
\label{tab:cost_analysis}
\footnotesize
\begin{adjustbox}{width=\columnwidth,center}
\begin{tabular}{@{}lccc@{}}
\toprule
\textbf{LLM Provider} & \textbf{Cost per Analysis} & \textbf{Processing Time} & \textbf{Tokens Used} \\
\midrule
GPT-4o & \$0.49 & 8-12 min & 2,800-3,200 \\
Llama 3.1 70B & \$0.12 & 10-15 min & 2,600-3,000 \\
PHI-4 & \$0.007 & 15-25 min & 2,400-2,800 \\
Qwen2.5-72B & \$0.08 & 12-18 min & 2,700-3,100 \\
\bottomrule
\end{tabular}
\end{adjustbox}
\end{table}

\begin{table}[!h]
\centering
\caption{Impact of removing individual subagents from the AI Data Scientist. Results include accuracy, run time, and the number of features generated for each configuration.}
\label{tab:ablation_study}
\footnotesize
\begin{adjustbox}{width=\columnwidth,center}
\begin{tabular}{@{}lcccc@{}}
\toprule
\textbf{Configuration} & \textbf{Churn Acc.} & \textbf{Diamond RMSE} & \textbf{Run Time} & \textbf{Features} \\
\midrule
Full Agent & 86.69\% & \$1,247 & 12 min & 147 \\
\makecell[l]{No Hypothesis\\Subagent}& 84.12\% & \$1,458 & 8 min & 89 \\
\makecell[l]{No Feature Engineering\\Subagent} & 83.45\% & \$1,592 & 10 min & 52 \\
Preprocessing Only & 82.01\% & \$1,734 & 6 min & 23 \\
\bottomrule
\end{tabular}
\end{adjustbox}
\end{table}

\section{Conclusion}
\label{sec:conclusion}

At its core, this work introduces a different kind of engine for data science, one that runs on hypotheses. By focusing not just on predictions but also on the questions that lead to them, our system reframes automated analysis as a process of exploration and structured reasoning, not merely optimization.

Throughout both experiments and real-world case studies, we see how hypothesis-powered agents can uncover meaningful patterns, offer clearer explanations, and help bridge the gap between statistical analysis and practical decision-making. Rather than simply aiming for accuracy, this approach prioritizes insights that are easier to interpret and apply.

More importantly, it points toward a different way forward for AI in analytics. Instead of replacing human judgment, it creates space for it. The system automates repetitive tasks while keeping the reasoning process open, transparent, and ready for expert input. As automation tools continue to evolve, we believe this focus on hypotheses, curiosity, careful testing, and structured inquiry offers a strong foundation for future advances in data science.

\bibliography{wileyNJD_AMA}   

\section*{Author contributions}
F.A. conceived the original idea and led system design. M.S.N. developed the hypothesis generation framework and conducted experimental evaluations. Z.I. implemented the feature engineering components and performed case study analysis. K.I. provided guidance on natural language processing components and business translation features. M.T. supervised the overall project and provided theoretical foundations. All authors contributed to writing and manuscript preparation.

\section*{Acknowledgments}
We thank Mohamed bin Zayed University of Artificial Intelligence for computational resources and funding support. We acknowledge our banking industry partner for providing real-world data and validation opportunities. We also thank the anonymous reviewers for their constructive feedback that improved the clarity and rigor of this work.

\section*{Conflict of interest}
The authors declare no competing financial interests or conflicts of interest that could influence this research.

\section*{Data availability}
Experimental datasets used in this study are publicly available through Kaggle. Banking case study data cannot be shared due to confidentiality agreements, but anonymized summary statistics are provided in the supplementary materials.

\section*{Code availability}
Implementation code and documentation will be made available upon publication to support reproducibility and further research.

\end{document}